# Wheel Impact Test by Deep Learning: Prediction of Location and Magnitude of Maximum Stress

Seungyeon Shin[1,a], Ah-hyeon Jin[1,a], Soyoung Yoo[1,3], Sunghee Lee[3],

ChangGon Kim[2], Sungpil Heo[2],

Namwoo Kang[1,3,*]

[1]Cho Chun Shik Graduate School of Mobility, KAIST, 34051, Daejeon, South Korea

[2]Hyundai Motor Company, 445706, Hwaseong-Si, Gyeonggi-Do, South Korea

[3]Narnia Labs, 34051, Daejeon, South Korea

[*]Corresponding author: nwkang@kaist.ac.kr

[a] Contributed equally to this work.

**Abstract**

For ensuring vehicle safety, the impact performance of wheels during wheel development must be ensured through a wheel impact test. However, manufacturing and testing a real wheel requires a significant time and money because developing an optimal wheel design requires numerous iterative processes to modify the wheel design and verify the safety performance. Accordingly, wheel impact tests have been replaced by computer simulations such as finite element analysis (FEA); however, it still incurs high computational costs for modeling and analysis, and requires FEA experts. In this study, we present an aluminum road wheel impact performance prediction model based on deep learning that replaces computationally expensive and time-consuming 3D FEA. For this purpose, 2D disk-view wheel image data, 3D wheel voxel data, and barrier mass values used for the wheel impact test were utilized as the inputs to predict the magnitude of the maximum von Mises stress, corresponding location, and the stress distribution of the 2D disk-view. The input data were first compressed into a latent space with a 3D convolutional variational autoencoder (cVAE) and 2D convolutional autoencoder (cAE). Subsequently, the fully connected layers were used to predict the impact performance, and a decoder was used to predict the stress distribution heatmap of the 2D disk-view. The proposed model can replace the impact test in the early wheel-development stage by predicting the impact performance in real-time and can be used without domain knowledge. The time required for the wheel development process can be reduced by using this mechanism.

# 1. Introduction

To ensure vehicle safety, vehicle wheels sufficiently durable to meet safety requirements need to be developed. Accordingly, a strict impact test must be performed during wheel development to test impact damage. However, completing a wheel design involves inspecting the wheel safety through the wheel impact test, which takes significant time and money owing to trial and error during wheel development. Therefore, vehicle manufacturing companies need a solution to reduce the time in the wheel design and manufacturing stages through a rapid impact analysis of the various design proposals.

Although the actual wheel impact test has been replaced by computer simulations such as finite element analysis (FEA), as proposed by Chang and Yang (2009), it is still time-consuming because the computationally expensive simulation process needs to be repeatedly executed, and FEA experts are required to inspect the wheel performance. Accordingly, recent studies have suggested methods for replacing FEA through deep learning methodologies in various applications such as stress distribution of the aorta (Liang et al., 2018), stress prediction for bottom-up SLA 3D printing processes (Khadilkar et al., 2019), stress prediction of arterial walls (Madani et al., 2019), stress field prediction of cantilevered structures (Nie et al., 2020), and natural frequency prediction of 2D wheel images (Yoo et al., 2021). These approaches can considerably contribute to accelerating the analysis process. However, there are still limitations to their use in real-world problems owing to the application domain. A more detailed review is presented in Section 2.

In this study, we present a 3D wheel impact performance prediction model based on deep learning that can replace 3D FEA of the aluminum road wheel impact test used in real-world product development processes. The objective of this study is to replace the 3D FEA process for wheel impact analysis, which requires high computational cost, to provide the impact performance of a wheel design in the conceptual design stage, thereby reducing the time required for wheel development. Synthetic 3D wheel data were generated through the 3D wheel CAD automation process (Oh et al., 2019; Yoo et al., 2021; Jang et al., 2022) using 2D disk-view images (spoke designs) and rim cross-sections, and the impact performance results were collected through FEA impact test simulation. Hence, we constructed a real-time prediction model that predicts the magnitude of the maximum von Mises stress, the corresponding location, and the overall stress distribution of the 2D disk-view using this mechanism.

The novelty of this study is as follows. This is the first study to apply deep learning to vehicle system impact tests. Second, various types of data were used as input and output through the multimodal autoencoder architecture to improve prediction accuracy. Third, to overcome the data shortage problem, a 3D convolutional variational autoencoder (cVAE) is used for transfer learning to extract important features of the 3D wheels. Through the proposed model, the impact performance of a wheel design can be checked in real-time, even in the conceptual design stage, by predicting the magnitude of the maximum von Mises stress, and the location of the maximum stress can also be known, informing the parts to be supplemented in the wheel design. The overall von Mises stress distribution of the 2D disk-view was also predicted, providing more information to the designer. Accordingly, this method can be easily utilized by general designers without engineering expertise, thereby enabling rapid impact performance inspection of various design proposals. The same process can be applied to any product that requires an impact test in addition to the wheels.

The remainder of this paper is organized as follows. Section 2 summarizes the related studies, and Section 3 presents the data collection and preprocessing steps for training the model and the architecture of the proposed model. The prediction results are discussed in Section 4. Finally, Section 5 presents conclusions, limitations, and directions for future work.

## 2. Related Work

Recently, studies to solve various real-world engineering problems using deep learning are being conducted, such as autonomous driving (Grigorescu et al., 2020), smart factories (Essien and Giannetti, 2020), and environmental engineering (Ostad-Ali-Askari et al., 2017; Ostad-Ali-Askari and Shayan, 2021). In addition, deep learning can be used in engineering problems to efficiently develop various products as the product development process requires repetitive iterations to obtain an optimal design (Kim et al., 2022; Shin et al., 2022; Yoo and Kang, 2021). In particular, design optimization for product development is time-consuming and computationally costly owing to repeated simulations such as FEA and computational fluid dynamics (CFD), which are essential for inspecting the safety of a product. Therefore, studies have recently emerged to replace the simulation process by applying various deep learning methodologies (Deng et al., 2020; Lee et al., 2020; Qian & Ye, 2021; Zheng et al., 2021).

This section focuses on deep learning studies that predict the stress distribution in structures. Liang et al. (2018), which is an early study that applied deep learning to replace FEA, predicted the aortic wall stress distribution according to the shape of thoracic aorta. Shape encoding of the aorta shape was performed through principal component analysis (PCA) for this purpose, and the stress distribution of the aorta was predicted through a neural network. Madani et al. (2019) predicted the maximum von Mises stress value and the corresponding location for 2D arterial cross-sectional images to replace finite element simulation using machine learning. Nie et al. (2020) predicted the stress distribution of a 2D linear elastic cantilevered structure subjected to a static load to accelerate structural analysis. In this study, two networks—SCSNet and StressNet—were proposed, and the von Mises stress distribution was predicted by inputting various structures, external forces, and displacement boundary conditions. However, the aforementioned studies are predictions for the 2D domain and have limitations when applied to actual product development. In real-world problems, high-dimensional data such as 3D data must be considered. However, high-dimensional data are difficult to train and require considerable training data. Therefore, appropriate data representation and training methods for high-dimensional data must be devised to replace 3D simulations with deep learning.

Similar to our study, Khadilkar et al. (2019) proposed two CNN-based networks to predict the stress distribution by layer for the bottom-up SLA printing process among manufacturing methods. In particular, the 2-stream CNN network, which exhibits the highest performance among them, uses a binary image of the cross-section and a 3D model up to the previous layer as input, in the form of a point cloud, to predict the stress distribution of a layer cross-section. The 2D image passes through the convolutional layer, and the 3D point cloud enters the network by adding each feature vector that passes through PointNet (Qi et al., 2017). This method has an input similar to that of our method. However, our proposed method predicts the maximum von Mises stress value and the corresponding location as well as the 2D stress distribution using voxel-based 3D data. Khadilkar et al. (2019) predicted the stress distribution for 2D domains, whereas our proposed methodology could predict the 3D coordinates of the maximum von Mises stress location for the 3D domain, enabling the replacement of the existing 3D FEA.

Two major issues must be considered for deep learning in the 3D domain. A considerable problem in the field of 3D deep learning is that it requires a large amount of training data. However, collecting a sufficient amount of 3D data in practice is difficult; therefore, we need to construct a model with high accuracy, even with limited data. Second, the representation of the 3D data is important when dealing with 3D data. In the field of 3D deep learning, representation methods such as point clouds, meshes, and voxels are commonly used.

The point-cloud-based method represents the shape through a set of points distributed near the surface of a 3D shape (Bello et al., 2020). However, the point cloud method has the disadvantage that expressing the details of a shape is difficult because it is sparse. The mesh-based method represents a 3D shape using a polygon-shaped face made of vertices. However, this method is sensitive to the quality of the input mesh and the surface patch of the shape may not be stitched. The voxel-based method uses volumetric data to express a 3D shape in a cube form. However, the voxel method requires a large amount of memory storage because it expresses the occupied and non-occupied parts (Ahmed et al., 2018). The resolution of the voxel needs to be increased to express the 3D shape in more detail, but the problem is that the higher the resolution, the more the parameters increase. Several studies have been conducted to solve the computational cost problem of the voxel method. For example, some studies have been performed to express 3D shapes with high resolution using an octree-based method (Häne et al., 2017; Riegler et al., 2017; Tatarchenko et al., 2017).

In this study, we propose a multimodal autoencoder architecture that uses multiple modalities in parallel as input and output. As in Bachmann et al. (2022), it is typically used to handle multiple sources such as images, text, and audio. We constructed a prediction model based on a multimodal autoencoder architecture that uses various dimensions of inputs and outputs in parallel to overcome the data shortage problem and reduce the computational cost by utilizing a latent vector of the input to reduce the dimension of the high-dimensional data. To this end, we used voxel-based 3D wheel data to extract the features of the 3D CAD data using a 3D CNN-based convolutional variational autoencoder (cVAE). Training was carried out using the latent vector of 3D CAD data and 2D wheel image data through the pretrained 3D cVAE model and the 2D convolutional autoencoder (cAE) model, and accurate results were derived even with 2,501 3D CAD data.

## 3. Deep Learning Framework for Wheel Impact Test

### 3.1. Overall Framework

The entire study process comprises four steps. In Stage 1, the 3D road wheel CAD datasets were automatically generated. Spoke designs were first collected from the Internet, provided by Hyundai Motors, and then generated using topology optimization. The rim cross-sections of the 3D wheels were also collected, and six representative designs were selected for use in 3D CAD generation. In Stage 2, wheel impact analysis was performed using the generated 3D wheel data. Based on the analysis results, post-processing was performed to remove outliers, and the magnitude of the maximum von Mises stress and its location coordinates were extracted. Stage 3 involves developing 3D cVAE and 2D cAE, which are dimensionality reduction models used to improve the performance of the proposed model. These models were used to reduce the dimensions of the input data. Finally, Stage 4 is the phase of developing a deep learning model that predicts the magnitude of the maximum von Mises stress, the corresponding location coordinates (x, y, and z), and the overall von Mises stress distribution in the 2D disk-view. The proposed model architecture and hyperparameters were selected by conducting various experiments while varying the input and output of the model. Finally, the performance of the proposed model was confirmed by conducting transfer learning using an actual wheel used in real life. The overall process proposed in this study is illustrated in Figure 1.

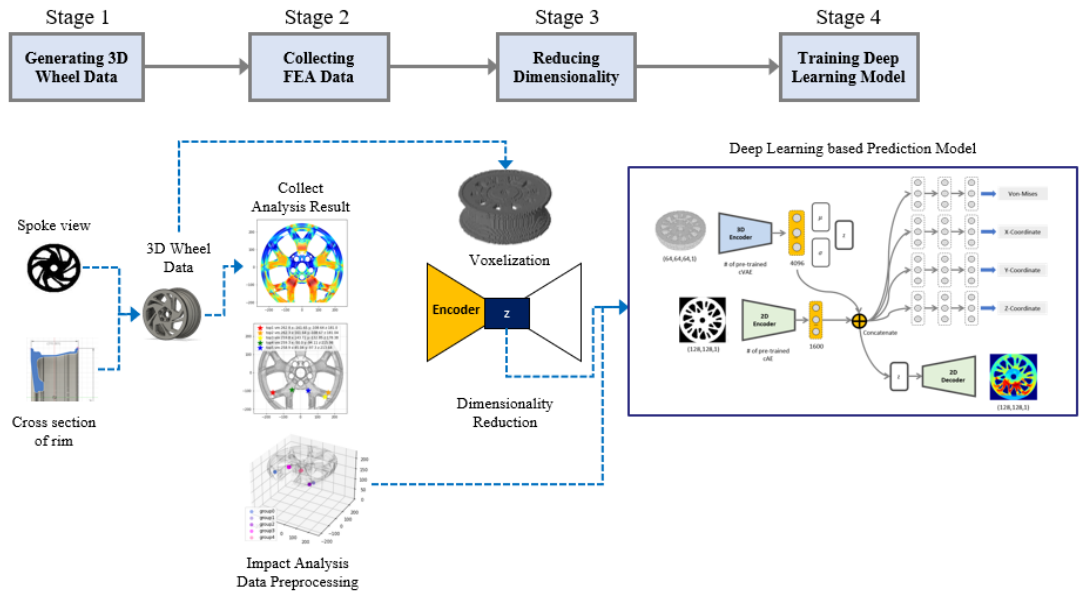

**Figure 1. Overall framework of the proposed method**

## 3.2. Stage 1: Generating 3D Wheel Data

Several detailed 3D wheel models used in reality are needed to construct an accurate wheel impact performance prediction model. However, considerable synthetic concept wheel data was generated and used for training because sufficiently detailed 3D wheel data are difficult to obtain. The 3D CAD automation framework proposed by Yoo et al. (2021) was used herein. The framework comprises a stage that handles 2D spoke designs (disk-view images) and rim cross-sections and a stage that creates these into 3D CAD. This process automatically generated a large amount of 3D roadwheel CAD. First, 2D work dealing with spoke designs and rim cross-sectional images was performed. Accordingly, spoke designs and rim-cross-section images were collected for this purpose, as explained in Sections 3.2.1 and 3.2.2.

### 3.2.1 Disk-view spoke design data collection and preprocessing

First, 2D disk-view spoke design images for 3D CAD generation are collected in various ways. The three main collection methods were as follows. First, 603 binary wheel images available on the Internet were collected, and topology optimization using the collected images as the reference design was performed to collect 177 generative design piece wheels. Topology optimization was performed on the wheel pieces, as shown in Figure 2, and they were rotated to make a complete wheel, solving the conventional problem of generative design wherein the symmetry of the generation result is not guaranteed. Here, 10 types of equal wheel pieces from 4 to 13 pieces were used, and generative design was performed by diversifying the similarity condition, load ratio condition, and volume ratio conditions among the topology optimization conditions. Additionally, 93 2D wheel images and 142 3D wheels from Hyundai Motors were converted into binary images. Thus, 1,015 2D disk-view spoke design images with 128 × 128 pixels were collected.

The collected spoke designs were post-processed using a series of processes, as shown in Figure 2. First, the edges of the anti-aliased 128 × 128 image are detected and saved using the Sobel operator. Among the edge values, the hub hole, which is the center hole of the wheel, and the edge of the outermost wheel were deleted, leaving only the edge values forming the spoke design. The coordinates of the remaining edge points are saved in a .csv file. A set of points forming a single loop is grouped together to use randomly aligned points for CAD modeling. When the Euclidean distance between points is smaller than a certain threshold, it is clustered in the same group. The points were evenly deleted to lower the density of the points in the sketch and draw a smooth spline. The preprocessed spoke design images are shown in Figure 3.

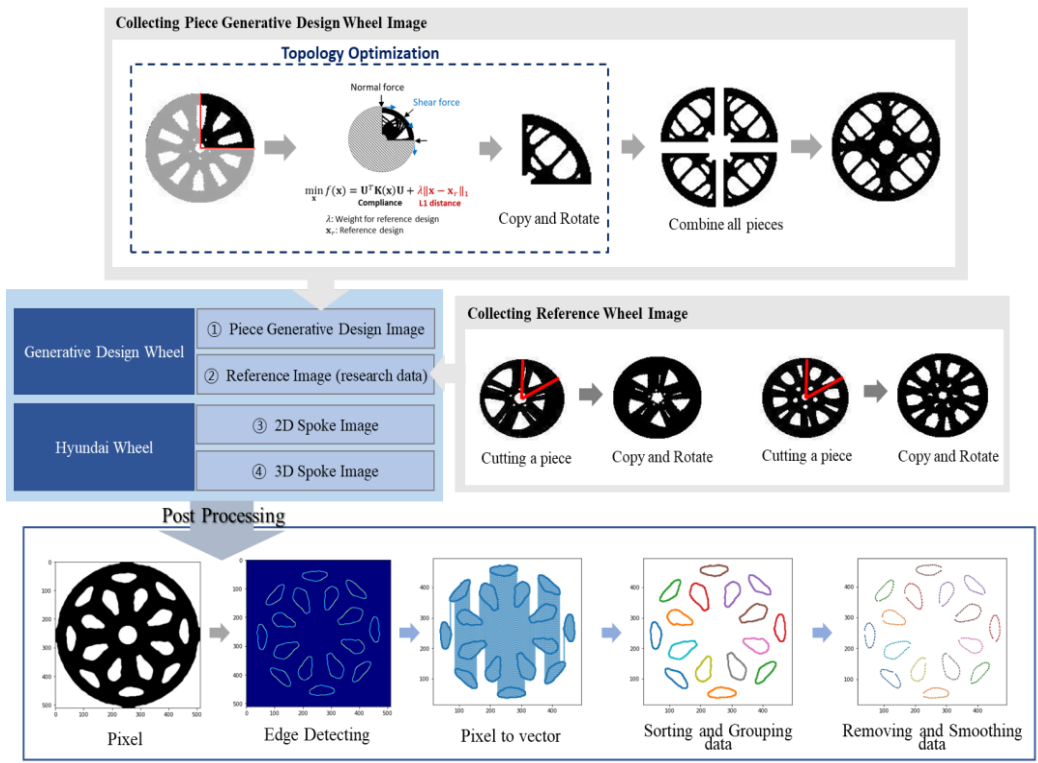

**Figure 2. Process of collecting spoke design data**

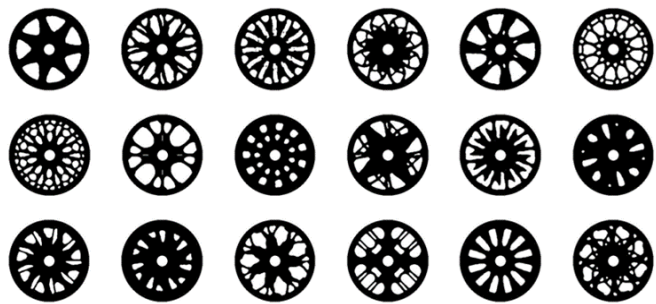

**Figure 3. Samples of the 2D disk-view spoke design data**

### 3.2.2 Rim cross-sectional data collection and preprocessing

In contrast to the spoke design images, several representative cross-sections were selected for the rim cross-section and combined with the spoke design images to create 3D wheels. The following section describes the dimension reduction for selecting representative rim cross-sections and preprocessing the images. First, the rim cross-sections of the wheels were collected at various angles to select representative rim cross-sections. Consequently, 2D rim cross-sectional images were obtained, as shown in Figure 4. Next, the images were cropped to split the lower part of the rim to remove unnecessary information and to concentrate on the upper part of the rim body.

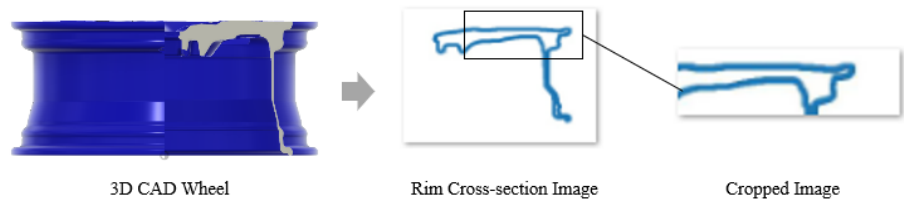

**Figure 4. Process of collecting rim cross-sectional data**

Thereafter, 70×235 pixel data were trained with a cAE to reduce the features of the rim cross sections into 64 dimensions, and they were clustered into six groups using K-means clustering and the elbow method. Six representative rim cross-sections, as shown in Figure 5, were selected, which were located at the center point of each cluster. Accordingly, we simplified the CAD automation while maintaining the diversity of rim cross-sections. The representative rim cross-sections selected through the process above undergo the same preprocessing steps as the spoke design images.

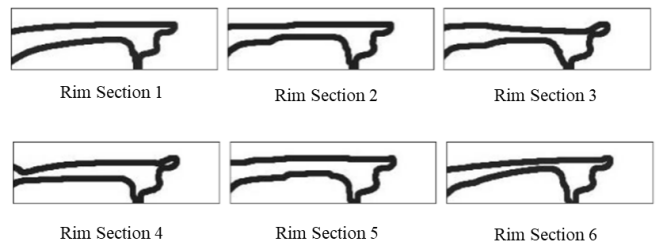

**Figure 5. Six representative rim cross-sectional data**

In addition, the coordinates of the spoke and lower parts of the rim were stored separately to ensure that the bodies could be individually created during modeling.

### 3.2.3 3D CAD automation

The following is a 3D CAD modeling process using the 2D disk-view spoke design images and the rim cross-sections prepared above. Autodesk's Fusion 360 Application Programming Interface tool (Autodesk, 2022) was used for 3D wheel CAD automation. The 3D modeling was automated by drawing lines and splines in a .csv file containing the image coordinates. The 3D CAD generation procedure is illustrated in Figure 6. First, after sketching the coordinates corresponding to the spoke part of the rim, a spoke body is created by revolving the sketch. Next, the coordinates for the disk-view spoke design were sketched and extruded, and predefined lug holes were sketched and extruded through the spoke body. Finally, after sketching the coordinates for the lower part of the rim section, the rim body was created by revolving the sketch and combining it with the spoke body to create the final wheel body and the CAD was then extracted. This process was automatically performed until the desired number of wheels was created.

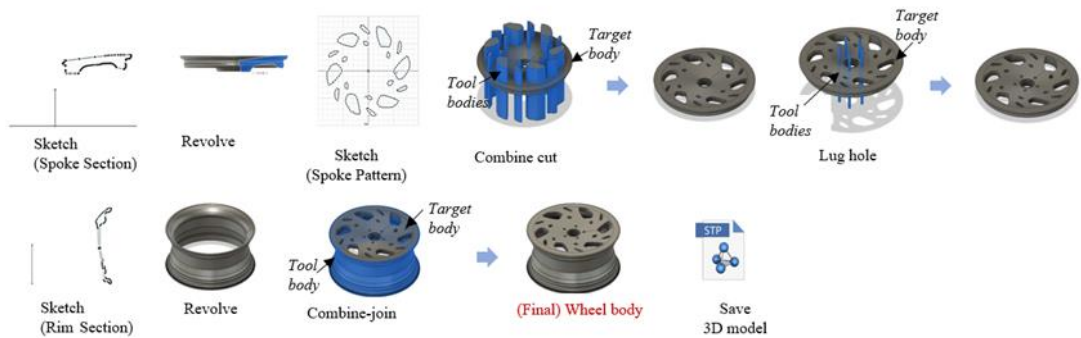

**Figure 6. 3D CAD modeling automation process**

## 3.3 Stage 2: Collecting FEA Data

### 3.3.1 Analysis Result Collection and Preprocessing

The 3D wheel model created by the 3D CAD automation process was subjected to wheel impact analysis to collect the von Mises stress response and preprocessed for use as training data. The collection of the analysis results and preprocessing were conducted in five steps, as shown in Figure 7.

First, a barrier mass value, which is the impact load used for the impact analysis, was assigned to the CAD file name of the generated CAD model. The barrier mass applied in the impact analysis was subjected to a force with a load unit (kg) at the same location. In this study, loads from 498.0 kg to 558.0 kg were divided into 1,000 equal intervals and randomly applied to ensure that the loads are not biased.

An inputdeck file containing information on the analysis condition, including the impact location and constraint conditions, was created for each CAD file using an automation program. The impact location, where the barrier mass is applied, was designated as the air hole of the wheel, which is located at the center of the widest hole among the wheel spoke holes. Wheel Impact Analysis, which is an impact analysis automation program provided by Altair based on HyperWorks (Altair HyperWorks, 2020), was used to automatically generate the inputdeck files. As a condition applied to the impact analysis automation program, the result types were set to element stresses (2D and 3D) and von Mises stresses, and the hub and bolt holes were constrained. In addition, the area near the air hole, which was the impact location, was set as the 3D elastic area.

The impact analysis was performed using Altair Optistruct (Altair Optistruct, 2020). The analysis was automated using Python-based code to continue the analysis for 2,501 files. After the analysis was completed, each result was extracted as a. csv file to be used as the label data for training. This process was performed through Wheel Impact Analysis, as previously mentioned. In the extracted .csv file, information about all the node IDs, xyz coordinates, von Mises stress values, and maximum principal stress values was recorded.

Before the location of the maximum von Mises stress was extracted from the analysis result file, the nodes with values that did not meet our criterion were removed. The criteria were as follows. First, the nodes near the impact location and the nodes at the lower part of the rim were excluded. In addition, the nodes with negative maximum principal stress values were excluded. Accordingly, stress in the compressive direction was excluded. The remaining nodes were arranged in descending order based on von Mises stress values. This process was performed using MATLAB (MATLAB 2020).

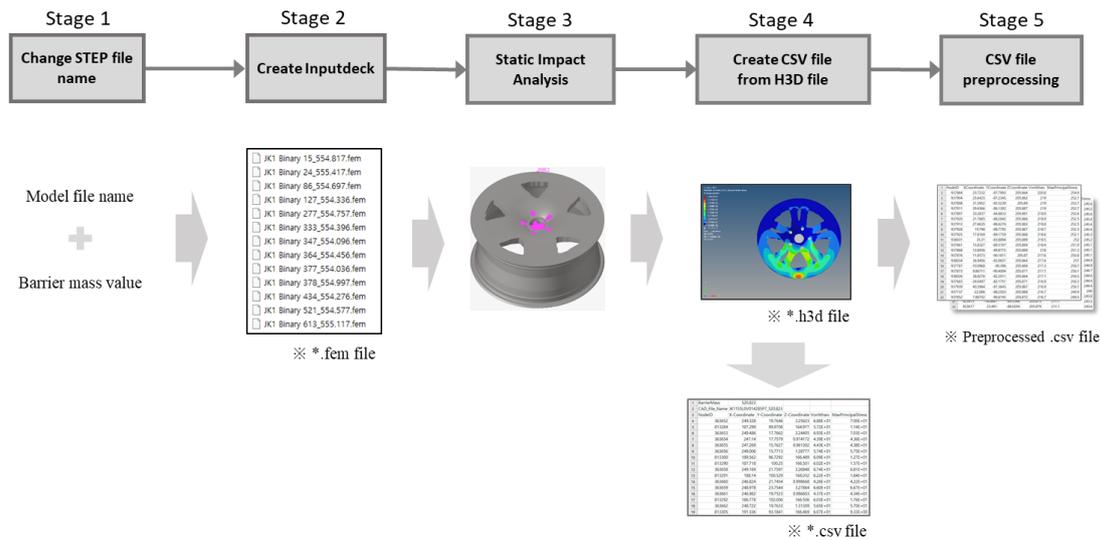

**Figure 7. Process of preprocessing wheel impact analysis result data**

### 3.3.2 Label Data Preprocessing

The label data preprocessing used for training was as follows. Based on the analysis results obtained in Section 3.3.1, the maximum von Mises stress value, corresponding coordinates, and overall stress distribution in the 2D disk-view were used as labels. The maximum stress values were obtained after removing outliers to accurately predict the maximum von Mises stress value and its location. The maximum stress point can be determined in three steps, as shown in Figure 8.

First, according to the preprocessed analysis results, the top 50 points were obtained based on the von Mises stress values. Next, the nodes with a 3D L2 distance of less than 10 mm from the location of maximum stress were grouped into one to cluster the high-stress-concentrated areas with the top 50 nodes. After excluding nodes that have already been grouped, a cluster with high stress concentration can be identified by repeating the same process for the remaining nodes. Following the selection of the maximum stress point from the top cluster, the coordinates of the node and von Mises stress values were stored and used as labels for training.

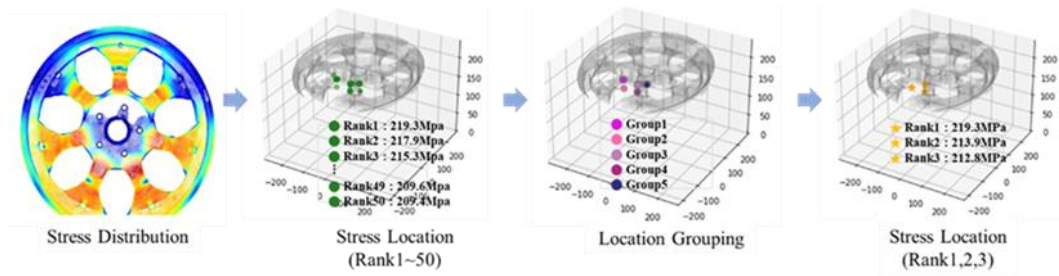

**Figure 8. Process of obtaining the maximum von Mises stress location**

## 3.4 Stage 3: Reducing Dimensionality

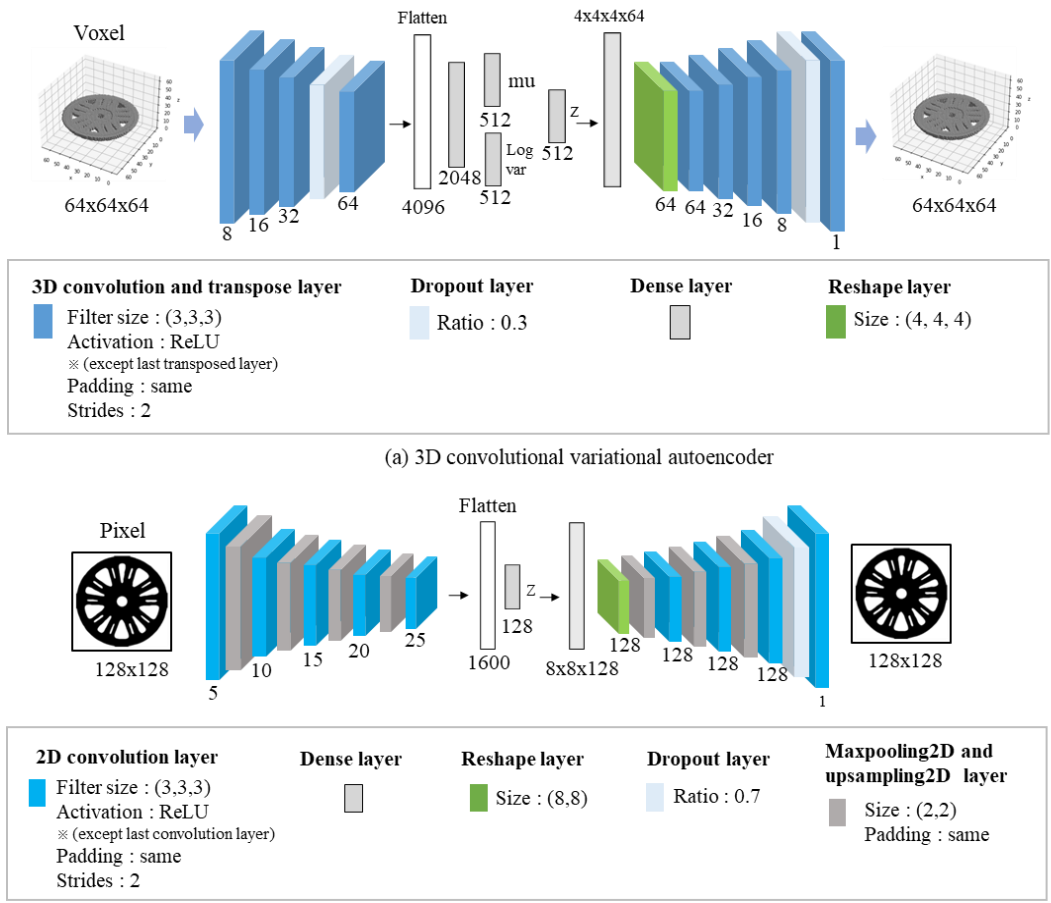

**Figure 9. Dimensionality reduction**

**(a) 3D convolutional variational autoencoder; (b) 2D convolutional autoencoder**

To learn the features of the wheels, two pretrained encoders were used to predict the impact performance of the wheels. The two encoders are described in this section (Figure 9). The features of the 2D disk-view spoke images were compressed using a 2D cAE. cAE is a network that converts unsupervised learning into a form of supervised learning by using the input as label data; the label data is trained to ensure that the difference between the input and output passed through the decoder is small. Consequently, the latent vector from the encoder compressed the features of the training data. The input pixel value of the 2D cAE was $128 \times 128$, and only the encoder from the trained cAEs was used afterwards. In addition, the 3D convolutional variational autoencoder (cVAE) is a network that learns to represent inputs as probability values based on probability distributions. This tool is primarily trained to use a decoder. An encoder was used in this study. The input voxel data have dimensions of $64 \times 64 \times 64$, and the latent vector that compresses the features of the input is used to construct the proposed prediction model, which is explained in Section 3.6.

## 3.5 Stage 4: Training Deep Learning Model

### 3.5.1 Data

The input data for the proposed model comprised 2D spoke design data, 3D voxel data, and barrier mass value used for the impact analysis. The labeled data comprise the maximum von Mises stress value, its coordinates, and a heatmap of the 2D disk-view stress distribution. A total of 2,501 3D road wheel CAD data were used, of which 1,753 (70%) were used as the training set, 374 (15%) as the validation set, and 374 (15%) as the test set.

First, the 2D spoke design data and 3D voxel data used as inputs were explained. From the original 3D CAD wheel model, 3D voxels and 2D images were individually extracted. 3D data are high-dimensional data and training them is difficult owing to the nature of high-dimensional data. However, the computational cost is also high. Accordingly, the lower part of the rim body was excluded, and only the spoke body was used to concentrate on the spoke body of the wheel, as shown in Figure 10 (a). This approach was possible because all wheels had the same shape of the lower rim body, and even in the actual wheel impact test, the stress occurring at the lower part of the rim body was not considered. Therefore, only the cropped 3D spoke body is converted to a voxel to be used as the 3D input, and the same assumption is applied to the detailed wheels in Section 4.3. The 3D CAD data with only the spoke body of the wheel was converted to $64 \times 64 \times 64$ voxels, covering approximately 7–8 mm per voxel. The 2D spoke design images corresponding to each 3D CAD dataset were represented by $128 \times 128$ pixels. The two types of wheel data are presented in Figure 10.

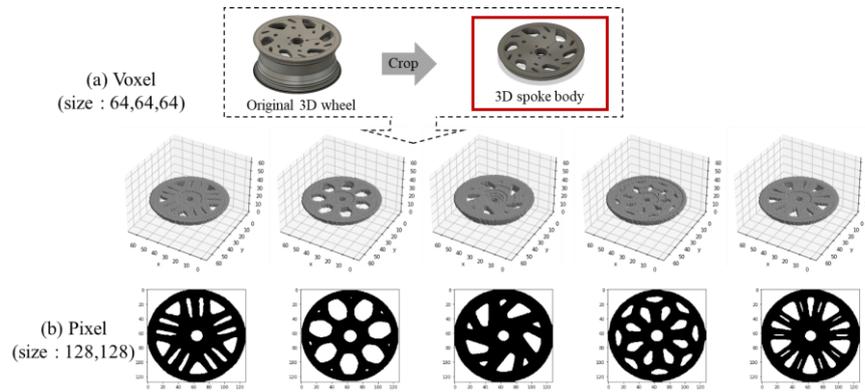

**Figure 10. Input wheel data**

**(a) 3D voxel wheel data (64 × 64 × 64); (b) 2D pixel wheel image data (128 × 128)**

Subsequently, the maximum von Mises stress values, the corresponding coordinates, and the 2D disk-view stress distribution heatmap, which are used as labels, are explained. A histogram of the coordinates of the wheel data and the maximum von Mises stress values is shown in Figure 11. Given that the coordinates and stress values had different distributions and ranges, min-max scaling was applied for both values, allowing for more stable training.

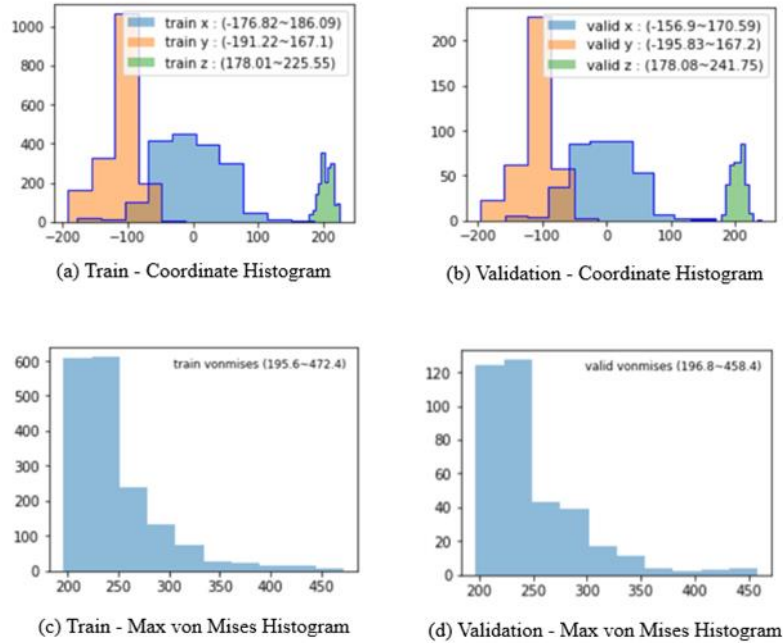

**Figure 11. Histogram of the coordinates and magnitude of maximum von Mises stress**

**(a) Train dataset coordinate histogram; (b) validation dataset coordinate histogram; (c) train dataset magnitude of the maximum von Mises stress histogram; (d) validation dataset magnitude of maximum von Mises stress histogram**

A 2D disk-view stress distribution heatmap was used as the label. This approach allows the proposed model to learn not only the maximum von Mises stress value and its location but also the overall von Mises stress distribution of the 2D disk-view. The von Mises stress values were mapped to a 128 × 128 matrix based on the analysis results obtained in Section 2.2 to create the corresponding heatmap data. A visualized image of the heat map is shown in Figure 12. The red color indicates the part with a high von Mises stress value, and the blue color denotes the portion with a low von Mises stress value.

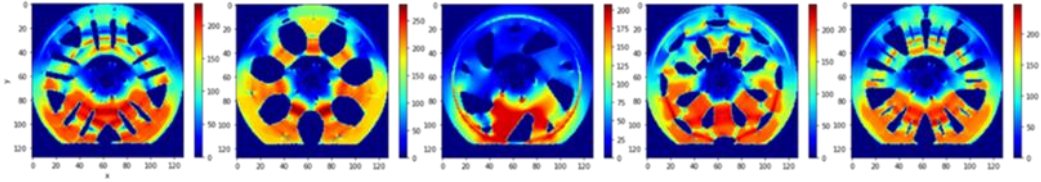

**Figure 12. Stress distribution heatmap of 2D disk-view**

### 3.5.2 Model Architecture

The architecture of the proposed model is described in this section. As shown in Figure 13, the entire network is configured to predict the maximum von Mises stress value, location, and heatmap by inputting a 3D voxel spoke body, 2D spoke design image, and barrier mass value. The model was built using the TensorFlow and Keras software.

To improve the training performance, the 2D wheel image and high-dimensional 3D voxel data received as inputs were used by extracting features through the pretrained 2D cAE and 3D cVAE described in Section 3.4. Through this, 3D data with dimensions of 64 × 64 × 64 were reduced to 512 dimensions using cVAE, and 2D data with dimensions of 128 × 128 were reduced to 128 dimensions using cAE. These latent vectors of the 2D wheel image and 3D voxel were combined with the barrier mass value; thus, a total of 641-dimensional input was fed into each regression model and decoder model. The proposed network comprises five models to predict the maximum von Mises stress value, corresponding xyz coordinates, and stress distribution in the 2D disk-view. First, a regression model composed of fully connected layers is used to predict the maximum von Mises stress coordinates and stress values. A rectified linear unit (ReLU) was used for the activation function, except for the last layer. A linear activation function is used for the final layer. A decoder composed of 2D convolutional layers and 2D upsampling is used to predict the heatmap. A rectified linear unit activation function was used, except for the last layer, and a linear activation function was used for the last layer. The architecture of each model is shown in detail in Figure 13. The loss function based on the mean squared error (MSE) used for training is shown in Equation (1).

$$Loss_{total} = \frac{1}{n}\sum_{i=1}^{n}(x_i - \hat{x}_i)^2 + \frac{1}{n}\sum_{i=1}^{n}(y_i - \hat{y}_i)^2 + \frac{1}{n}\sum_{i=1}^{n}(z_i - \hat{z}_i)^2 + \frac{1}{n}\sum_{i=1}^{n}(s - \hat{s}_i)^2 + \frac{1}{n}\sum_{i=1}^{n}(I_i - \hat{I}_i)^2, \qquad (1)$$

where $n$ is the number of input data, $i$ is the i-th input data, $x_i$ is the i-th ground truth x-coordinate, $\hat{x}_i$ is the i-th predicted x-coordinate, $y_i$ is the i-th ground truth y-coordinate, $\hat{y}_i$ is the i-th predicted y-coordinate, $z_i$ is the i-th ground truth z-coordinate, $\hat{z}_i$ is the i-th predicted z-coordinate, $s_i$ is the i-th ground truth von Mises stress value, $\hat{s}_i$ is the i-th predicted von Mises stress value, $I_i$ is the i-th ground truth heatmap, and $\hat{I}_i$ is the i-th predicted heatmap. The Adam Optimizer was used for training, with a learning rate of 0.0001, batch size of 32, and 1,000 epochs. Bayesian optimization-based and random search-based hyperparameter searches were tested to increase the model accuracy. In conclusion, hyperparameters such as the learning rate, batch size, and epoch were selected through a random search.

The MSE loss for each epoch during the training process is shown in Figure 14. The training results converged well for both the training and validation sets.

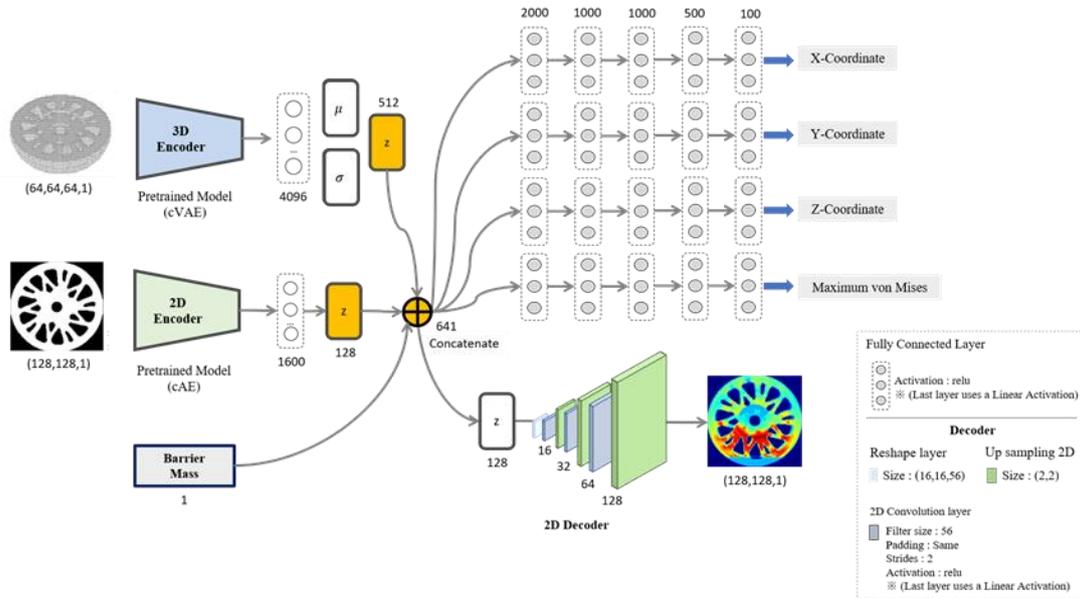

**Figure 13. Proposed model architecture**

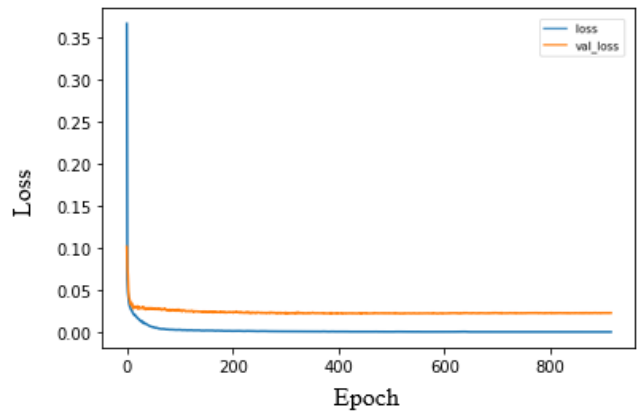

**Figure 14. MSE loss for each epoch in the training process**

## 4 Results and Discussion

In this section, we analyze the results of the proposed model. In Section 4.1, we define the various metrics used to evaluate the proposed model and discuss the prediction results. To visually show the results, we divided the cases into good and bad prediction cases to visualize the predicted location and the ground truth location on the 3D wheels and the predicted stress distribution heatmap. In Section 4.2, we compare several model architectures by varying the input and output, proving that the proposed model has the best model architecture. Finally, in Section 4.3, actual road wheel data were tested using transfer learning to demonstrate the scalability of our proposed model to the actual road wheel.

## 4.1 Proposed Model Prediction Result

Various metrics were used to evaluate the prediction results of the model. The mean and median 3D Euclidean distance errors were checked for the coordinates. Meanwhile, the mean absolute percentage error (MAPE) and Pearson correlation R-value were checked for von Mises stress values. The accuracy of the heatmap was evaluated using the root-mean-square error (RMSE). The equations for the 3D Euclidean distance error, MAPE, and RMSE were as follows:

$$Euclidean\ Distance\ Error\ _{3D} = \sqrt{(x_i - \hat{x}_i)^2 + (y_i - \hat{y}_i)^2 + (z_i - \hat{z}_i)^2}, \tag{2}$$

$$MAPE\ _{von\ Mises} = 100 \times \frac{1}{n} \sum_{i=1}^{n} \left|\frac{s_i - \hat{s}_i}{s_i}\right|, \tag{3}$$

$$RMSE\ _{heatmap} = \sqrt{\frac{1}{n} \sum_{i=1}^{n} (I_i - \hat{I}_i)^2}. \tag{4}$$

The test results for the proposed model with the corresponding metric showed that the mean 3D Euclidean distance error was 31.49 mm, and the median 3D Euclidean distance error was 24.73 mm. Considering that the size of one voxel of the corresponding 3D CAD data was 7–8 mm, a meaningful prediction result was obtained. Figure 15 shows a histogram of the overall distribution of the 3D Euclidean error with respect to the predicted coordinates. The figure shows that the validation and test results had an error of 50 mm or less. Furthermore, the relative 3D Euclidean distance error was examined to compare the Euclidean distance error relative to the wheel size. The Euclidean distance error was normalized based on the wheel diameter (approximately 483 mm). The relative mean 3D Euclidean distance error was 6.52%, and the relative median 3D Euclidean distance error was 5.12%. The MAPE of the maximum von Mises stress value was 2.99%, and the R value was 0.96, indicating that an accurate prediction was possible for the test set.

Scatter plots of the predicted maximum von Mises stress values for the training, validation, and test sets are shown in Figure 16. The scatter plots show that the prediction accuracy was slightly low for wheels with a large maximum von Mises stress value. This result was obtained because the maximum von Mises stress value of most training datasets was less than 300 MPa, similar to that in Figure 11(c). If we additionally collect wheel data with a large maximum von Mises stress value to solve the data imbalance problem, better training results can be obtained. The RMSE for each pixel of the restored heatmap was 0.0490, indicating that the heatmap was meaningfully restored. The test results of the proposed model and the training and validation sets are presented in Table 1.

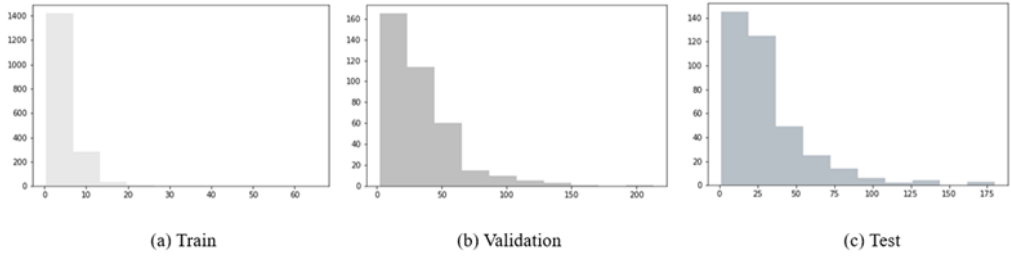

**Figure 15. Histogram of the coordinate 3D Euclidean distance error**

**Table 1. Proposed model prediction result**

| Dataset | 3D Euclidean Distance Error (Mean) (mm) | 3D Euclidean Distance Error (Median) (mm) | Relative 3D Distance Error (Mean) (%) | Relative 3D Distance Error (Median) (%) | Von Mises MAPE (%) | Von Mises (R) | Heatmap RMSE (MPa) |
|---|---|---|---|---|---|---|---|
| Train | 5.40 | 4.79 | 1.12 | 0.99 | 0.44 | 0.99 | 0.0332 |
| Validation | 34.23 | 26.97 | 7.09 | 5.58 | 2.87 | 0.96 | 0.0480 |
| Test | 31.49 | 24.73 | 6.52 | 5.12 | 2.99 | 0.96 | 0.0490 |

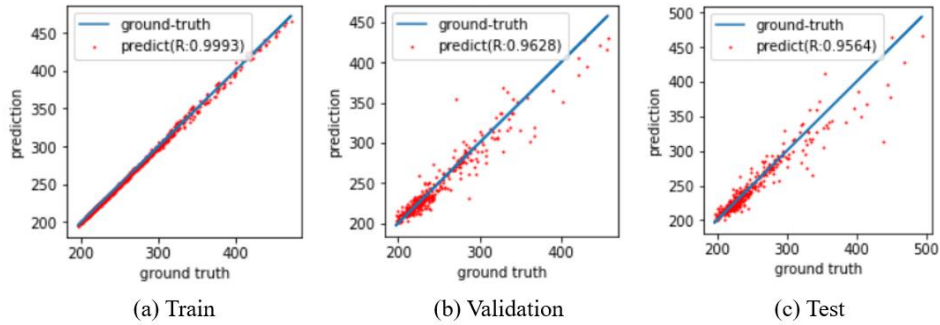

**Figure 16. Scatter plot of the model's prediction and the ground truth**

**(a) Scatter plot of the train dataset; (b) scatter plot of the validation dataset; (c) scatter plot of the test dataset**

As discussed above, the prediction accuracy of the maximum stress value and the stress distribution heatmap was high. Accordingly, a more detailed analysis of the predicted coordinates is conducted. The 1D Euclidean distance errors for the x, y, and z coordinates were examined. The result is the same as the above-mentioned 3D Euclidean distance error equation, with the only difference being that it is expressed in one dimension. In Table 3, the relative mean 1D Euclidean distance errors of the test set were 5.20%, 2.36%, and 1.32% for the x-, y-, and z-coordinates, respectively. The relative median 1D Euclidean distance errors of the test set were 3.72%, 1.36%, and 1.03% for the x-, y-, and z-coordinates, respectively. The Euclidean distance error results for each coordinate for the training, validation, and test sets are presented in Table 2. The results show that the prediction rate for the z-coordinate was relatively high. Given that the z-axis corresponds to the height of the wheel, it is expected to be relatively easy to predict. However, the prediction rate for the x-coordinate is the lowest, which is considered the most difficult coordinate because the wheel has a symmetrical design with respect to the impact location.

**Table 2. Proposed model 1D coordinate prediction result**

| Dataset | X Distance Error (Mean) (mm) | X Distance Error (Median) (mm) | Y Distance Error (Mean) (mm) | Y Distance Error (Median) (mm) | Z Distance Error (Mean) (mm) | Z Distance Error (Median) (mm) |
|---|---|---|---|---|---|---|
| Train | 3.43 | 2.60 | 2.14 | 1.75 | 2.34 | 2.06 |
| Validation | 27.32 | 20.64 | 12.28 | 6.78 | 7.05 | 5.48 |
| Test | 25.14 | 17.98 | 11.38 | 6.55 | 6.38 | 4.97 |

**Table 3. Proposed model 1D coordinate prediction result (relative error)**

| Dataset | X Relative Distance Error (Mean) | X Relative Distance Error (Median) | Y Relative Distance Error (Mean) | Y Relative Distance Error (Median) | Z Relative Distance Error (Mean) | Z Relative Distance Error (Median) |
|---|---|---|---|---|---|---|
| Train | 0.71 | 0.54 | 0.44 | 0.36 | 0.48 | 0.43 |
| Validation | 5.66 | 4.27 | 2.54 | 1.40 | 1.46 | 1.14 |
| Test | 5.20 | 3.72 | 2.36 | 1.36 | 1.32 | 1.03 |

Finally, we visualize the prediction results of the proposed model. The visualization of the ground truth maximum von Mises stress location and the predicted location is shown in Figure 17. Figure 18 shows the rendered wheel data with the ground truth and the prediction location.

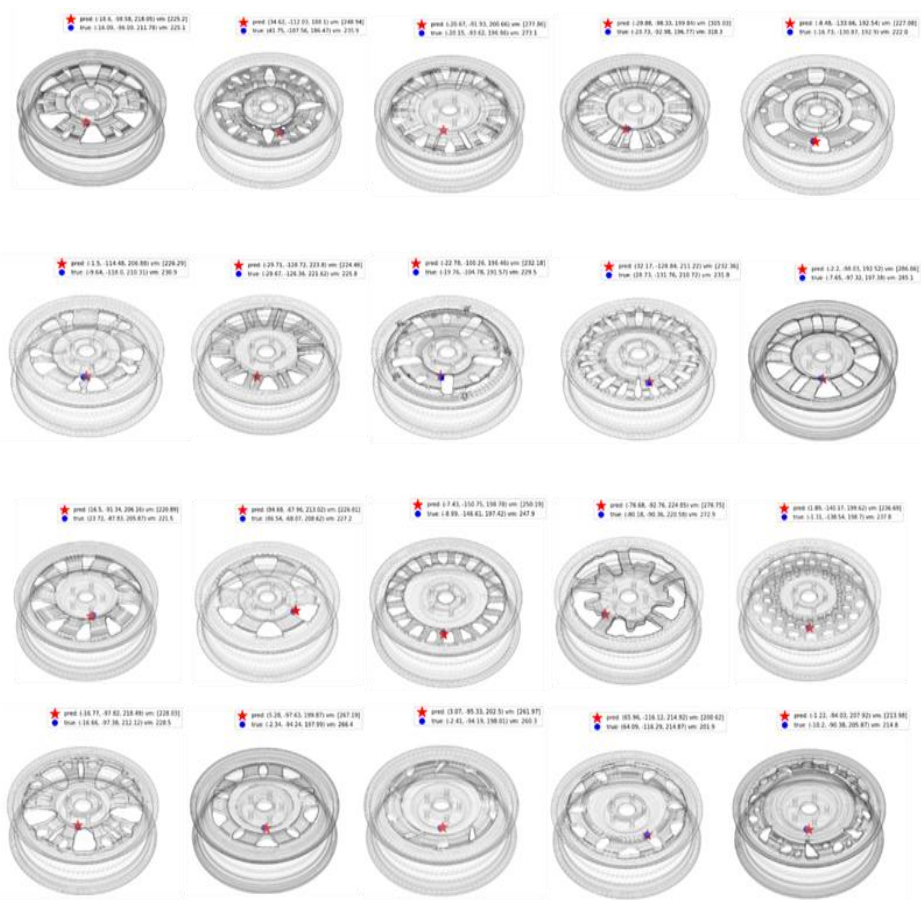

**Figure 17. Visualization of the prediction location (red star) and the ground truth location (blue dot)**

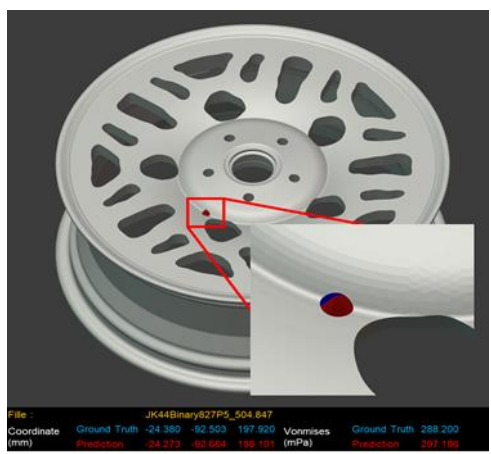

**Figure 18. Visualization of the rendered wheel and the prediction (ground truth [blue point] and prediction [red point])**

In addition, a comparison between the ground truth stress distribution heatmap and the predicted stress distribution heatmap is shown in Figure 19. The figure shows that the heatmap was meaningfully reconstructed.

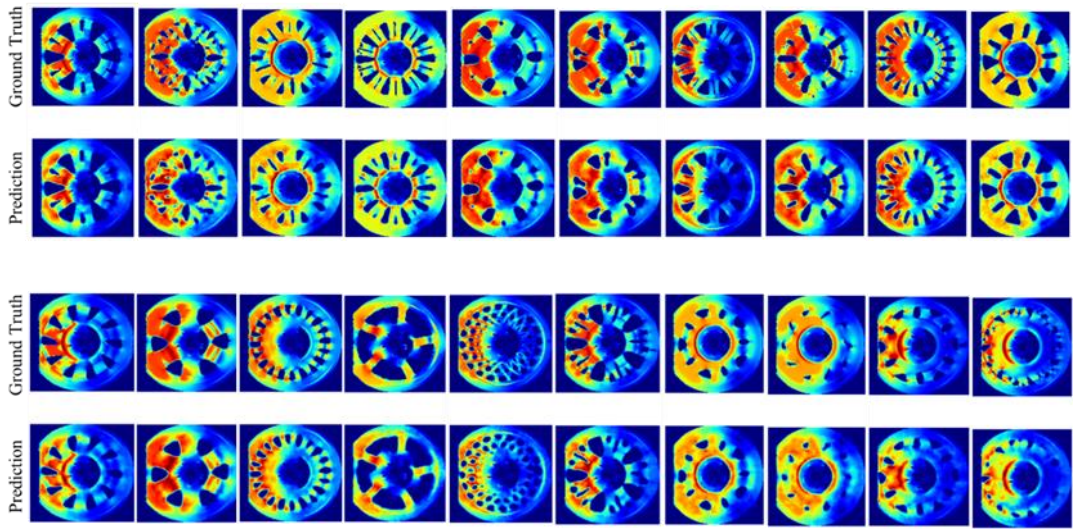

**Figure 19. Comparison between the ground truth and the prediction of the stress distribution heatmap**

Figure 20 shows the results for the poorly predicted case. Figure 20(a) shows a visualization of the predicted maximum stress location and the actual location. Figure 20(b) shows the ground truth and prediction heatmap corresponding to the wheel shown in Figure 20(a). The results showed that most poorly predicted cases were predicted to be locations where high stress of a similar size occurred near the actual maximum stress location. Given that many locations have high stress of similar size, users should determine the vulnerable area by not only looking at the location of the maximum stress and its value but also the overall stress distribution to make an accurate judgement.

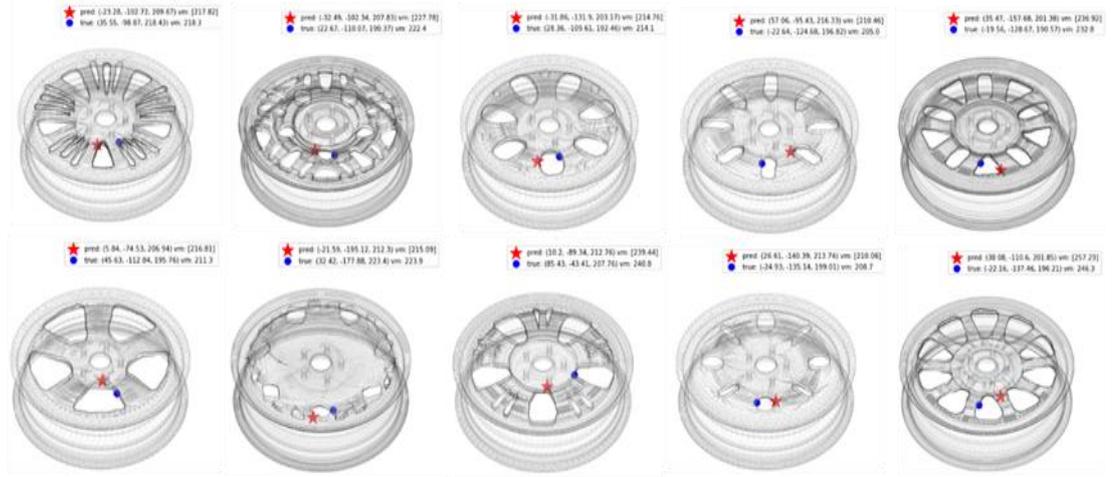

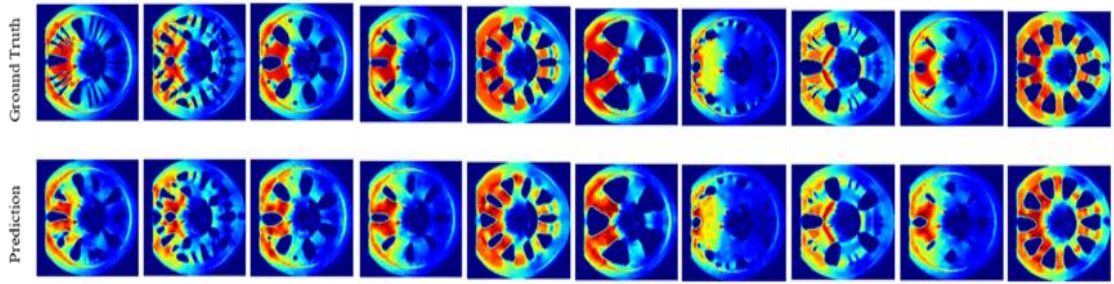

**Figure 20. Visualization of the bad prediction cases**

**(a) Ground truth and prediction location; (b) ground truth and prediction of the stress distribution heatmap**

### 4.2 Architecture Comparison

As shown in Section 3.5, the model used various inputs in parallel, and the maximum von Mises stress value, coordinates, and overall stress distribution of the 2D disk-view can be predicted with high accuracy through this mechanism.

The prediction accuracy may decrease with an increase in the number of variables to be predicted, because the training weight differs for each variable to be predicted during the training process. We examined how additionally predicting the stress distribution heatmap of the 2D disk-view affects the model performance by comparing the prediction performance of Model B, which does not predict the stress distribution heatmap, and the proposed model. Model B has the same architecture and hyperparameters as the proposed model, except for the decoder. The results are as follows. In the case of Model B, for the test set results, the 3D mean relative Euclidean distance error was 6.92%, 3D median relative Euclidean distance error was 5.47%, and MAPE for the maximum von Mises stress value was 3.07%, as shown in Table 5. However, the 3D mean relative Euclidean distance error was 6.52%, the

3D median relative Euclidean distance error was 5.12%, and the MAPE for the maximum von Mises stress value was 2.99% for the proposed model. Therefore, the proposed model can predict the 2D disk-view stress distribution while maintaining high prediction performance for the maximum stress value and coordinates.

Furthermore, we compared the prediction results of Model A, which does not use a 2D wheel image as the input, and Model B under the same condition that the stress distribution heatmap of the 2D disk-view is not restored to examine the performance change of the model according to the input type. Model A was trained with the same architecture and hyperparameters as Model B, except for the input type. The comparison in Tables 4 and 5 confirms that the method using 3D wheel data and 2D wheel images together is more effective in predicting the maximum von Mises stress value. A comparison of the structures of Models A and B and the proposed model is shown in Figure 21.

**Table 4. Different models' validation dataset prediction result**

| Model | 3D Euclidean Distance Error (Mean) (mm) | 3D Euclidean Distance Error (Median) (mm) | Relative 3D Distance Error (Mean) (%) | Relative 3D Distance Error (Median) (%) | Von Mises MAPE (%) | Von Mises (R) |
|---|---|---|---|---|---|---|
| Model A | 34.64 | 27.08 | 7.17 | 5.61 | 3.69 | 0.93 |
| Model B | 34.21 | 27.93 | 7.08 | 5.78 | 2.81 | 0.97 |
| Proposed Model | 34.23 | 26.97 | 7.09 | 5.58 | 2.87 | 0.96 |

**Table 5. Different models' test dataset prediction result**

| Model | 3D Euclidean Distance Error (Mean) (mm) | 3D Euclidean Distance Error (Median) (mm) | Relative 3D Distance Error (Mean) (%) | Relative 3D Distance Error (Median) (%) | Von Mises MAPE (%) | Von Mises (R) |
|---|---|---|---|---|---|---|
| Model A | 33.53 | 25.08 | 6.94 | 5.19 | 3.88 | 0.93 |
| Model B | 33.42 | 26.42 | 6.92 | 5.47 | 3.07 | 0.95 |
| Proposed Model | 31.49 | 24.73 | 6.52 | 5.12 | 2.99 | 0.96 |

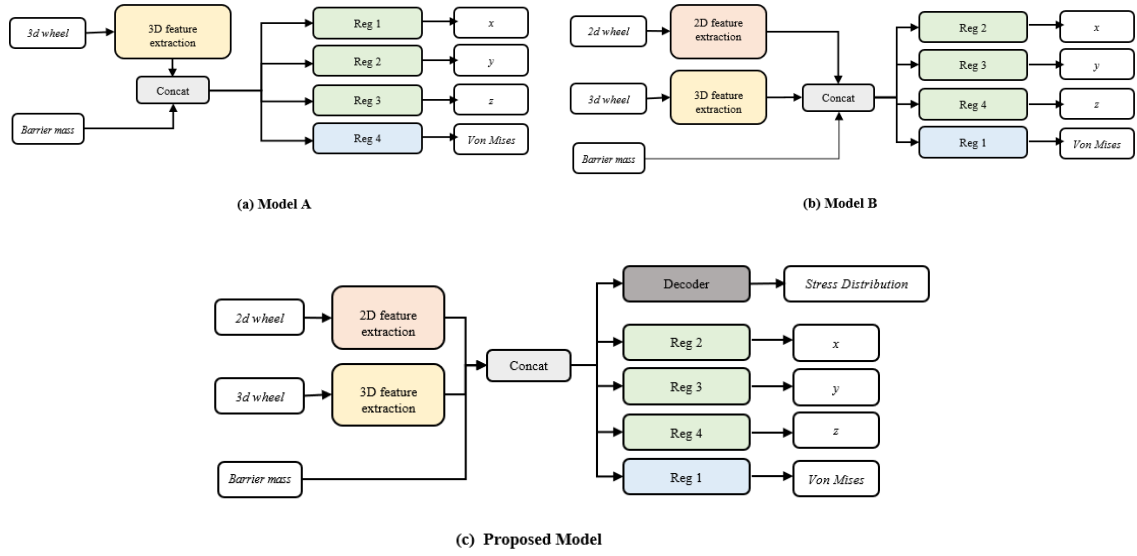

**Figure 21. Comparison between the proposed model and other models**

**(a) Architecture of Model A; (b) architecture of Model B; (c) architecture of the proposed model**

### 4.3 Different Domain Data

We checked the prediction performance of the proposed model using 63 detailed wheel data (used in real vehicles) provided by Hyundai Motors to examine the applicability of the proposed method in actual product development, which is the ultimate goal of our research. Figure 22 shows the distance between the concept wheels and detailed wheels in the latent space. We embedded the concept and detailed wheels into a 2D latent space and visualized them using T-SNE (Van der Maaten and Hinton, 2008). The figure shows that the distance between the data is large and that the concept wheels are clustered by their rim type. Owing to the dissimilarity between these data, we employed transfer learning to apply the proposed model to 63 detailed wheel datasets. Among them, 53 detailed wheel data points were used to train the transfer learning model, and 10 detailed wheel data points were used to evaluate the results. Based on the proposed architecture shown in Figure 13, all layers except for the last layer of the regression models and decoder model were frozen. All other training conditions remained the same as those in the proposed model, and only the learning rate was reduced to 0.00001.

Table 6 shows that the relative mean 3D Euclidean distance error was 6.11%, relative median 3D Euclidean distance error was 5.05%, MAPE of the maximum von Mises stress value was 5.94%, and RMSE for the restored heatmap was 0.1636.

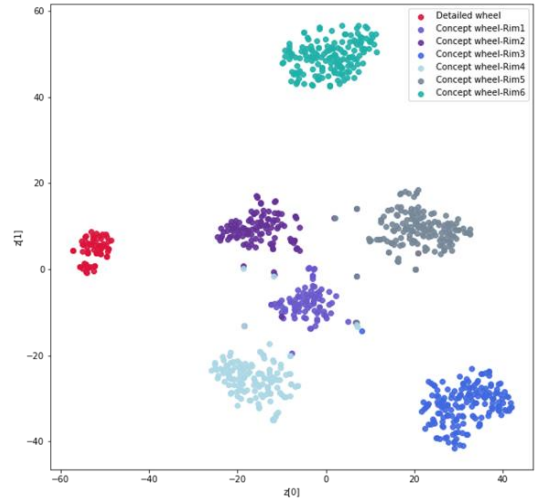

**Figure 22. Latent space of concept wheel and detailed wheel using T-SNE**

The results of the relative 3D Euclidean distance error show that the vulnerable location of a wheel was accurately predicted by looking at the wheel features, even though the data domains were different. However, the prediction results for the maximum stress value are inferior. Given that the detailed wheel data have a detailed design, such as intaglio (curved surface), it is quite different from that of the training data, and the distributions of the x- and y-coordinates of the wheel data are distinct from those of the training dataset, i.e., the concept wheels. Most of the predicted maximum von Mises stress values were predicted to be lower than the actual stress values, and the predicted heatmap in Figure 24(b) was also predicted to have a lower overall value. This problem is owing to the maximum von Mises stress value of the detailed wheel, which is inevitably higher than that of the concept wheel for a similar spoke design owing to the detailed design. As shown in Figure 23, which shows the volume distribution of the two domains, the average volume of the detailed wheel was 4.62e6 mm$^3$, and the average volume of the concept wheel was 5.78e6 mm$^3$. This finding indicates that the volume of the detailed wheel is relatively small. Thus, the maximum von Mises stress value of the detailed wheel is bound to be larger. Therefore, additional research is required to predict the maximum von Mises stress value in the actual wheel data domain due to this problem. However, the location of the maximum stress was well predicted.

**Table 6. Proposed model detailed wheel dataset prediction result (transfer learning)**

| Dataset | 3D Euclidean Distance Error (Mean) (mm) | 3D Euclidean Distance Error (Median) (mm) | Relative 3D Distance Error (Mean) (%) | Relative 3D Distance Error (Median) (%) | Von Mises MAPE (%) | Von Mises (R) | Heatmap RMSE (MPa) |
|---|---|---|---|---|---|---|---|
| Test | 21.69 | 16.64 | 4.49 | 3.45 | 4.10 | 0.65 | 0.1590 |

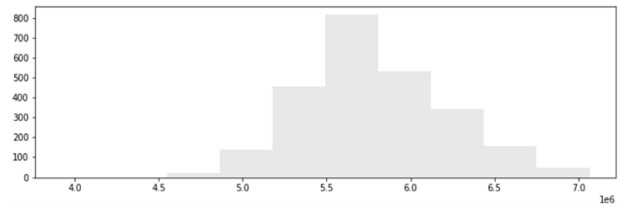

(a) Concept wheel volume histogram

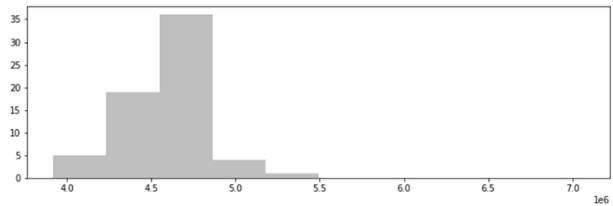

(b) Detailed wheel volume histogram

**Figure 23. Comparison between the concept wheel data volume and the detailed wheel data volume**

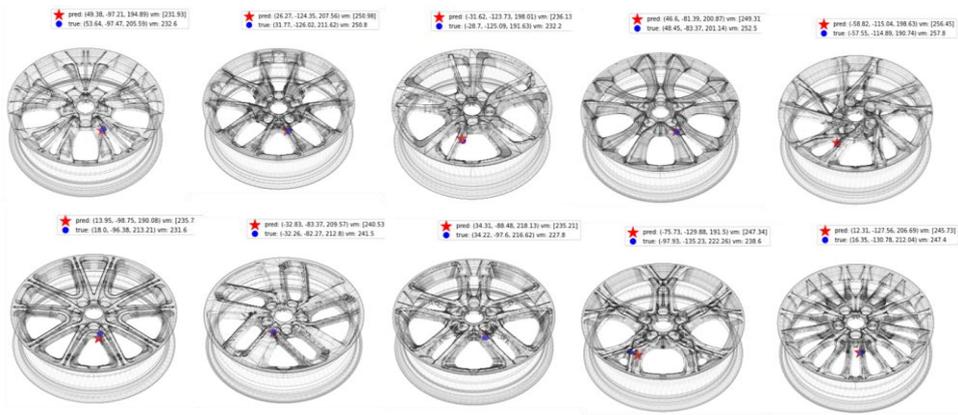

(a)

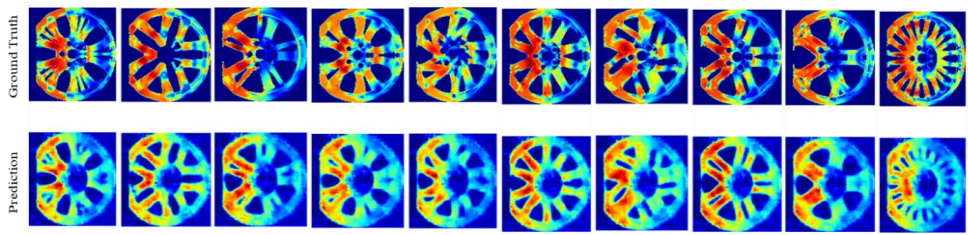

(b)

**Figure 24. Visualization of the prediction for a detailed wheel dataset**

**(a) Ground truth and prediction location; (b) ground truth and prediction of the stress distribution heatmap**

The proposed model took approximately 27.41 min to train with one GPU (Geforce RTX 3090). Meanwhile, the prediction of the impact performance of the wheel data using the trained model was only approximately 0.001 s. This finding shows that this method is highly effective compared with conventional impact analysis such as FEA, which requires an average of 1 h or more (CPU 64 CORE) to analyze the wheel data. Therefore, we proved that our methodology is suitable for accelerating the 3D FEA process and reducing computational cost.

# 5 Conclusion

In this study, we proposed a real-time wheel impact performance prediction model based on deep learning to replace the time-consuming 3D FEA for wheel impact tests used in the real-world engineering industry. Using the generated concept wheels to construct a prediction model, we were able to expand the proposed model to actual road wheels using transfer learning. For this purpose, 3D concept wheels were generated, and the wheel impact performance was collected through FEA. The 3D and 2D wheel data were then compressed into the latent space and inputted in parallel to train the prediction model. As a result, we obtained a 3D wheel impact performance prediction model that predicts the magnitude of the maximum von Mises stress, corresponding location, and stress distribution of the 2D disk-view.

The proposed model can play a role in helping designers to quickly derive an optimal design without engineering knowledge by providing a quick prediction about the impact performance, even with a conceptual design at the design phase. In particular, this study shows the applicability of this method to actual product development because we applied this method to 3D synthetic road wheel data. Considering the characteristics of 3D CAD data, which make it difficult to collect a large amount of high-dimensional data, we derived an accurate prediction model by using latent vectors containing the features of input data through pretrained autoencoder-based models. Consequently, a prediction model with high accuracy was constructed with only a total of approximately 2,501 3D CAD data. The model constructed in this manner is meaningful because it replaces the 3D FEA and provides real-time impact analysis results of an unseen wheel design because not only the maximum von Mises stress value but also the location of the maximum stress occurrence and the stress distribution of the 2D disk-view can be predicted.

The contributions of this study are summarized as follows. To the best of our knowledge, this is the first study to apply 3D deep learning to vehicle system impact analysis. A 3D deep learning model capable of predicting the maximum stress value, coordinates, and stress distribution was presented. Accordingly, the impact analysis results for a new design can be obtained in real-time using this mechanism.

Second, we proposed a multimodal autoencoder architecture that used various types of data in parallel as the input and output. This architecture had a structure wherein a 3D voxel, 2D image, and scalar value were input in parallel. Meanwhile, the 2D images were restored in parallel with a vector value as the output. The proposed architecture was verified through an ablation study.

Third, the data shortage problem was overcome through transfer learning using a 3D convolutional variational autoencoder (cVAE). After extracting features by reducing the dimensions through 3D VAE for input data, the encoder was used during transfer learning for supervised learning because the 3D data required a large amount of data to directly perform supervised learning.

However, this study had some limitations. First, the direct application of this model to the detailed design used in the actual product was limited as mentioned in Section 4.3. Accordingly, we would like to conduct further research using this model. Considering that obtaining a sufficient amount of detailed design data was difficult, we would like to conduct a future study to expand it to a detailed road wheel design by applying the domain adaptation of Ganin and Lempitsky (2015) based on the proposed model trained from the concept wheel design. In addition, the accuracy of the prediction model increased if more 3D wheel data were collected. Given that a large amount of 3D CAD data was difficult to collect, we would like to use a physics-informed neural network (Raisi et al., 2019; Raissi et al., 2020), which is based on physical information, to solve the data shortage problem and reduce unnecessary computational costs in the data collection process. Gu and Golub (2022) applied a physics-informed neural network to the 2D thin-walled structure problem. In future, we wish to conduct a study using physics information based on the method of calculating partial differential equations (PDEs) of the input wheel by reflecting the PDE in the loss function. This method is expected to increase the accuracy of the prediction model because the training is performed by utilizing a physics-informed loss function rather than simply predicting using a black box.

**Acknowledgements**

This work was supported by the Hyundai Motor Company, the National Research Foundation of Korea (2018R1A5A7025409), and the Ministry of Science and ICT of Korea (No.2022-0-00969 and No.2022-0-00986).